\documentclass[letterpaper, 10 pt, conference]{ieeeconf}  

\IEEEoverridecommandlockouts                              

\usepackage{cite}

\usepackage{graphicx} 
\usepackage{amsmath} 
\usepackage{amssymb}  

\DeclareMathOperator*{\argmin}{arg\,min}
\usepackage{subcaption}
\usepackage[font=footnotesize]{caption}
\usepackage{algorithm2e}
\usepackage[capitalise]{cleveref}
\usepackage[abbreviations]{glossaries-extra}
\glsdisablehyper

\usepackage{multirow}

\glssetcategoryattribute{acronym}{indexonlyfirst}{true}
\newabbreviation{rrt}{RRT}{Rapidly-exploring Random Tree}
\newabbreviation{rrg}{RRG}{Rapidly-exploring Random Graph}
\newabbreviation{rba}{RBA}{Rubber Band Algorithm}
\newabbreviation{irba}{IRBA}{Iterative Rubber Band Algorithm}
\newabbreviation{dp}{DP}{Dynamic Programming}
\newabbreviation{idp}{IDP}{Iterative Dynamic Programming}
\newabbreviation{toi}{ToI}{Target of Interest}
\newabbreviation{roi}{RoI}{Region of Interest}
\newabbreviation{poi}{PoI}{Pose of Interest}
\newabbreviation{tsp}{TSP}{Traveling Salesman Problem}
\newabbreviation{gtsp}{GTSP}{Generalized Traveling Salesman Problem}
\newabbreviation{tspn}{TSPN}{TSP with Neighborhoods}
\newabbreviation{gtspn}{GTSPN}{GTSP with Neighborhoods}
\newabbreviation{mtp}{MTP}{Multi-Goal Path Planning Problem}
\newabbreviation{mtpgr}{MTPGR}{MTP for Goal Regions}
\newabbreviation{minlp}{MINLP}{Mixed Integer Nonlinear Programming}
\newabbreviation{ga}{GA}{Genetic Algorithm}
\newabbreviation{hrkga}{HRKGA}{Hybrid Random-Key Generic Algorithm}
\newabbreviation{rsvp}{RSVP}{Rover Sequencing and Visualization Program}
\newabbreviation{esa}{ESA}{European Space Agency}
\newabbreviation{esric}{ESRIC}{European Space Resources Innovation Centre}
\newabbreviation{src}{SRC}{Space Resources Challenge}
\newabbreviation{tsdf}{TSDF}{Truncated Signed Distance Field}
\newabbreviation{tpp}{TPP}{Touring-a-sequence-of-Polygons Problem}
\newabbreviation{prm}{PRM}{Probabilistic Roadmap}
\newabbreviation{ompl}{OMPL}{Open Motion Planning Library}
\newabbreviation{art}{ART Planner}{ANYmal Rough Terrain Planner}

\usepackage{booktabs}

\title{\LARGE \bf
SMUG Planner: A Safe Multi-Goal Planner for Mobile Robots in Challenging Environments
}

\author{Changan Chen$^{1}$, Jonas Frey$^{1,2}$, Philip Arm$^{1}$, and Marco Hutter$^{1}$}

\begin{document}

\maketitle
\thispagestyle{empty}
\pagestyle{empty}

\begin{abstract}
Robotic exploration or monitoring missions require mobile robots to autonomously and safely navigate between multiple target locations in potentially challenging environments. Currently, this type of multi-goal mission often relies on humans designing a set of actions for the robot to follow in the form of a path or waypoints. In this work, we consider the multi-goal problem of visiting a set of pre-defined targets, each of which could be visited from multiple potential locations. To increase autonomy in these missions, we propose a safe multi-goal (SMUG) planner that generates an optimal motion path to visit those targets. To increase safety and efficiency, we propose a hierarchical state validity checking scheme, which leverages robot-specific traversability learned in simulation. We use LazyPRM* with an informed sampler to accelerate collision-free path generation. Our iterative dynamic programming algorithm enables the planner to generate a path visiting more than ten targets within seconds. Moreover, the proposed hierarchical state validity checking scheme reduces the planning time by $30\%$ compared to pure volumetric collision checking and increases safety by avoiding high-risk regions. We deploy the SMUG planner on the quadruped robot ANYmal and show its capability to guide the robot in multi-goal missions fully autonomously on rough terrain.
\end{abstract}

\section{INTRODUCTION}
Multi-goal missions are commonly seen in exploration, inspection, and monitoring scenarios. In those missions, a robot needs to visit numerous targets of interest for detailed investigation, for example to deploy instruments, collect samples, or read measurements from gauges. However, in current missions, the autonomy of the robot to navigate and plan is limited and relies on human operators~\cite{arches,esa,hvdc}. 



Space exploration is a common application of such multi-goal missions. One example is the geological mission I of the ARCHES analog demonstration mission~\cite{arches}. In this mission, a flying system and a rover map the environment and detect multiple targets of interest. A second rover collects samples and returns them to the landing system one at a time according to scientists' prioritization. A planner that generates a global path visiting all identified targets according to the map provided by other robots could increase the autonomy of the robot team. Another space exploration example is the \gls{src} launched by the \gls{esa} and the \gls{esric}~\cite{esa} that simulates a lunar prospecting mission, in which multiple targets need to be mapped and characterized. Several teams used a multi-robot strategy to map the environment and find the targets before starting a close-up investigation of the targets. However, none of the teams used a multi-goal planner: They either operated via waypoints provided by an operator or used heuristics to always visit the next closest target.

\begin{figure}[t!]
\centering
\includegraphics[width=0.48\textwidth]{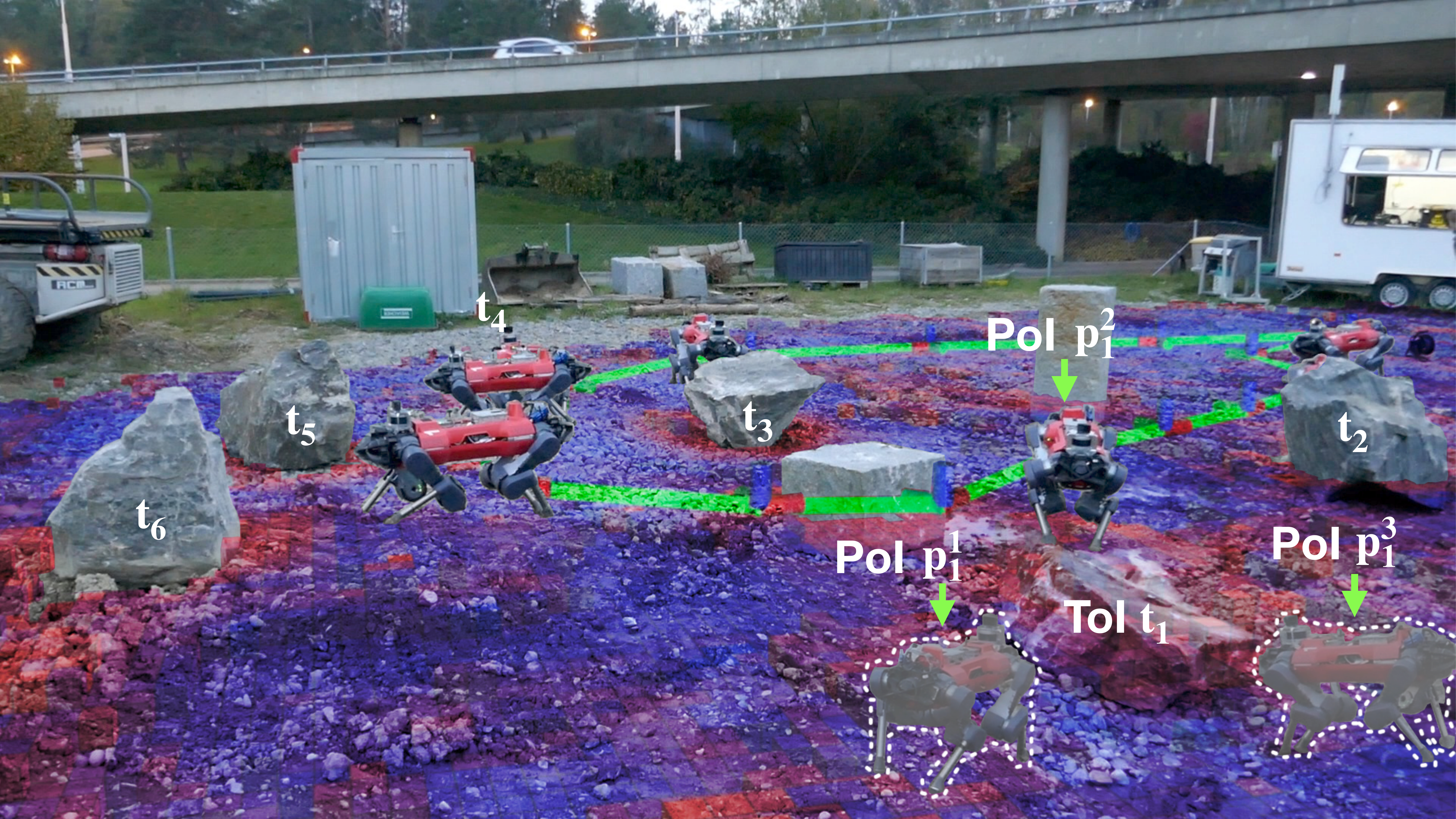}
\caption{The ANYmal quadruped robot autonomously performs a multi-goal mission of visiting six stones (Targets of Interest (ToIs):$t_i$) at one of the corresponding Pose of Interest (PoI): $p_i^j$ following the global path colored in green. The traversability map is overlaid on the real scene. The red regions are untraversable whereas the blue regions are traversable. The PoIs $p_1^1, p_1^2, p_1^3$ for the ToI $t_1$ are indicated by the green arrows. $p_1^2$ is selected, whereas the other two not selected PoIs are gray.}
\label{fig:intro}
\vspace{-0.6cm}

\end{figure}
Apart from exploration missions, multi-goal navigation is also relevant for industrial inspection and monitoring tasks. An example is the ARGOS (Autonomous Robot for Gas and Oil Sites) Challenge~\cite{argos} initiated by the company Total and ANR, in which the robots are required to navigate over a decommissioned gas dehydration skid to perform inspection tasks, such as reading sensor dials and valve positions, at various targets. Gehring et al.~\cite{hvdc} provide a solution to the industrial inspection of an offshore HVDC platform with the ANYmal quadruped robot. After the environment is mapped using the onboard sensors, the operator records the desired target location, at which detailed visual or thermal inspection needs to be done, and defines the via-point for the robot.

No planner exists that allows for fully autonomous multi-goal mission conduction. This results in such missions mostly relying on human operators, causing delays, sub-optimal decision-making, and limiting efficiency.

In this paper, we present SMUG planner, a multi-goal planner to plan a safe path to visit a set of targets at the poses in their vicinity for detailed investigation. We refer to the targets as \gls{toi} and the valid poses to visit them as \gls{poi}, as illustrated in~\cref{fig:intro}. More precisely, we present a planner to visit one \gls{poi} per \gls{toi}, which we refer to as multi-goal mission in this paper. This is the \gls{gtsp} with collision-free path planning. The large number of collision-free paths involved in the problem makes it impossible to simply deploy an off-the-shelf \gls{gtsp} solver, such as GLNS~\cite{glns}, which would require computing all collision-free paths and results in infeasible computational cost. How to efficiently ensure the safety and the optimality of the global path makes this an especially challenging problem.

SMUG planner implements a two-stage planning schema: At first the optimal visit sequence of the \gls{toi}s using a \gls{tsp}~\cite{tsp} solver is determined. Secondly, the optimal \gls{poi} to visit each \gls{toi} is determined using \gls{dp} in an iterative fashion, which we refer to as \gls{idp}. We use LazyPRM*~\cite{rrt*} with an informed sampler to efficiently generate the optimal collision-free paths. 
To guarantee safety, we propose a hierarchical obstacle avoidance strategy using a robot-specific traversability estimation module~\cite{lvn}.
This design allows for replanning during the mission and onboard deployment. SMUG planner can therefore guarantee autonomy in communication-denied environments, which are common in many exploration and monitoring scenarios.

Our contributions consist of the following:
\begin{itemize}
    \item A multi-goal path planner for mobile robots solving the GTSP with collision-free paths that can generate a safe path to visit more than ten targets in seconds.
    \item A novel obstacle avoidance scheme that reduces the planning time by 30\% compared to the pure volumetric collision checking and avoids high-risk regions.
    \item Hardware deployment on the ANYmal legged robot in a multi-goal mission on rough terrain. To the best of our knowledge, this is the first time a global planner is deployed to a real-world GTSP with collision-free path planning on a mobile robot. 
\end{itemize}

\section{RELATED WORK} 
Our work builds upon prior work developed in the context of sampling-based path planning and methods for solving the \gls{tsp} and its variants modeling multi-goal problems. Given our objective of real-world deployment, we specifically review existing global planners deployed on mobile robots.

\paragraph{Sampling-Based Path Planning} One family of popular path planning methods is sampling-based path planners~\cite{rapid,nbv,gbp,art}, which allows for a continuous state space instead of discretization as in grid-based methods. \gls{prm}~\cite{prm}, and its optimal version PRM*~\cite{rrt*} build a probabilistic roadmap for the environment and effectively reuse the information collected across different queries,  making these planners suitable for planning across multiple start-goal pairs in the same environment. Because the collision-checking of nodes and edges is costly and efficiency is important to our planner, LazyPRM*~\cite{lazyprm*} is well-suited, since it only checks edge validity if it may construct the optimal path, which reduces the planning time. Gammel~et~al.~\cite{informed} proposed to restrict the sampling space based on the cost of the best path found so far, thus accelerating path improvement. Our method builds upon LazyPRM* with an informed sampler to achieve efficient path planning in multi-goal missions.

\paragraph{Multi-Goal Sequencing Problem} 
\gls{tsp} is the problem of finding the optimal sequence visiting a set of targets. A \gls{tsp} instance considers each target as a single point. Its two variants, \gls{gtsp}~\cite{gtsp} and \gls{tspn}~\cite{tspn}, consider the targets as sets of points and continuous neighborhoods respectively. This approach is more accurate in many real-world scenarios, where a point often only needs to be approached instead of visited exactly. Exact algorithms proposed to solve \gls{gtsp} include using Branch and Cut~\cite{ant}, Lagrangian relaxation~\cite{lagrangian} or transforming \gls{gtsp} to an equivalent \gls{tsp} instance~\cite{transform}. Other works solving \gls{gtsp} and \gls{tspn} based on heuristics such as the ant-colony optimization~\cite{antcolony}, genetic algorithm~\cite{rkga,hrkga} and Lin-Kernighan heuristic~\cite{lk} or formulate the problem as \gls{minlp}~\cite{minlp}. However, these works ignore the computationally expensive collision-free path planning for a mobile robot navigating in complex environments. 
%
%
Alatartsev et al.~\cite{onoptimizing} divide the \gls{tspn} into two subproblems: Given the visiting point of each target, find the optimal visiting sequence. And given the visiting sequence, find the optimal visiting points. These subproblems are solved iteratively. Although the obstacle avoidance necessary in a real-world scenario is ignored, the strategy of splitting the \gls{tspn} into two subproblems is insightful, and we adopt it in our method.
Gentilini~\cite{hrkga} focuses on combinatorial optimization with \gls{hrkga} while assuming all path costs are known a priori. However, the author provides two cases in simulation that consider obstacle avoidance, which is achieved by planning the required collision-free path for each chromosome in the genetic algorithm, making it expensive and thus unsuitable for our problem that requires the fast online generation of safe paths. 

The variants of \gls{gtsp} and \gls{tspn} addressing collision-free path planning are called \gls{mtp}~\cite{mtp} and \gls{mtpgr}~\cite{mtpgr} respectively. Gao et al.~\cite{automatic} divide the \gls{mtpgr} into two subproblems as in~\cite{onoptimizing}. However, instead of iterating between them, they first solve for the optimal sequence once. Then, the path between any two points is initially assumed to be the straight line, and refined by alternating iteratively using the \gls{rba}~\cite{rba} to find the visiting point and planning the needed path accordingly. Thus, collision-free paths are only planned if they may construct the optimal global path, largely reducing the number of paths to plan. However, the \gls{rba} is prone to locally optimal path cost, since it adjusts the visiting point solely based on the two neighboring paths. Moreover, the authors only illustrate their method on a 2D grid map and lack real-world demonstration. 

Instead of solving \gls{mtpgr}, we model the multi-goal problem as an \gls{mtp} by assuming the \gls{toi}s to be discrete sets of \gls{poi}s rather than a continuous neighborhood, because one may want to deploy an instrument to the \gls{toi}s, which requires a certain angle of attack or impose other constraints, resulting in disconnected valid \gls{poi}s. However, the most general choice is to model each \gls{toi} as a set of disconnected neighborhoods, which results in \gls{gtspn}~\cite{gtspn}. Our assumption of each \gls{toi} being a discrete set simplifies the problem, while still allowing for multiple valid \gls{poi} proposals.

\paragraph{Planners for Mobile Robots}
Multiple existing planners for mobile robots have been successfully deployed in the real world~\cite{gbp,gbp2,tare,nbv}. Although they are not designed for the multi-goal mission, we use previously proven concepts and design patterns.

Several works adopted a bifurcated global and local planning structure~\cite{gbp,gbp2,tare}, where the local planner has detailed information in the vicinity of the robot, handles obstacle avoidance, and is guided by a global planner with coarser paths. Following this structure, we generate the entire traversing path and use it to guide a local planner that refines the path locally at a higher frequency.


A discrete volumetric map is often used to represent the environment~\cite{gbp,gbp2,nbv}, which can be obtained by Voxblox~\cite{voxblox}. In \cite{gbp} \cite{nbv}, and \cite{gbp2}, the authors consider the volumetric collision with the environment, however, not accounting for terrain traversability, which is crucial for the feasibility and safety of the path followed by a local planner. Therefore, we propose to leverage an in simulation-trained legged robot-specific traversability module~\cite{lvn}, within the global planner, to evaluate the robot-specific terrain traversability.



\section{PROBLEM STATEMENT}
We aim to solve the problem of finding the safe global path that visits every \gls{toi} in multi-goal missions. We formulate the problem as an \gls{mtp}. The robot can visit a \gls{toi} at multiple \gls{poi}s in a neighborhood of it. In this work, we limit the problem to visiting one \gls{poi} per \gls{toi}. 

Let $\mathcal{C}$ be the state space representing the environment and $\mathcal{C}_{collision}$ the set consisting of the states in $\mathcal{C}$ that result in a collision between the robot and environment. The collision-free space is therefore $\mathcal{C}_{free}=\mathcal{C}\setminus \mathcal{C}_{collision}$. A path is denoted as a discrete representation through a set of waypoints, i.e. $W(w_0,w_n)=\left\{w_0, w_1, w_2, …, w_n\right\}$ with $w_i\in\mathcal{C}$. $W$ represents the path starting at state $w_0$, traversing through $w_1$, $w_2$ ..., and ending at $w_n$. The cost of a path segment $\{w_i, w_{i+1}\}$ is given by a positive definite function $c(\cdot,\cdot): \mathcal{C}\times \mathcal{C} \mapsto \mathbb{R}_{\geq 0}$. The total cost of the path $W$ is the sum of the cost of all its path segments, i.e.: 
\begin{equation}
    Cost(W(w_0,w_n)) = \sum_{i=0}^{n-1}{c(w_i, w_{i+1})}.
\end{equation}
The input and output of the problem are summarized as:

\textbf{Input}: 
\begin{itemize}
    \item Robot start and end pose $s\in\mathcal{C}_{free}$.
    \item A set of \gls{toi} poses $\mathcal{T}=\{t_i|t_i\in\mathcal{C}, i\in\{1,\cdots,n\}\}$ with $t_i\in\mathcal{C}$.
    \item For each \gls{toi} $t_i$, a set of $n_i$ \gls{poi} $\mathcal{P}_i = \{p_i^j|p_i^j\in\mathcal{C}_{free}, j\in\{1,\cdots,m_i\}\}$.
\end{itemize}

\textbf{Output}: A collision-free path $W(w_0,w_n) = \left\{w_0,...,w_n\right\}$ with minimum cost visiting every \gls{toi} $t_i\in \mathcal{T}$ at a \gls{poi} $p_i^j \in \mathcal{P}_i$, starting at $s$ and ending at the same state:
\begin{equation}
\begin{aligned}
\min_{W} \quad & Cost(W)\\
\textrm{s.t.} \quad & p_i^j\in W, \forall{t_i\in\mathcal{T}},\exists{p_i^j\in\mathcal{P}_i}, \\
& w_0 = w_n = s, \\
& w_i\in\mathcal{C}_{free}, \forall w_i \in W
\end{aligned}
\end{equation}
The constraint $w_n = s$ can be relaxed but is used here to facilitate the notation.
In this work, the state space $\mathcal{C}$ is chosen to be $SE(2)$, suitable for ground robots. A state $s$is thus expressed by its 2D coordinates $x_s, y_s$ and yaw $\psi_s$ as
\begin{equation}
    s = 
    \begin{bmatrix}
        x_s, y_s, \psi_s 
    \end{bmatrix}^\top.
\label{eq:s}
\end{equation}
For the cost function $c(\cdot,\cdot): \mathcal{C}\times \mathcal{C} \mapsto \mathbb{R}_{\geq 0}$, we use the \gls{ompl}~\cite{ompl} default distance function in $SE(2)$:
\begin{equation}
\label{eq:cost}
    c(s_1, s_2) = w_t\left\|
    \begin{bmatrix}
        x_{s_1} \\
        y_{s_1}
    \end{bmatrix} -
    \begin{bmatrix}
        x_{s_2} \\
        y_{s_2}
    \end{bmatrix}\right\|_2^2 + w_r d(\psi_{s_1}, \psi_{s_2}),
\end{equation}
where $w_t, w_r \in \mathbb{R}^+$ are the weights for the translational and the rotational cost, respectively, and $d(\cdot,\cdot): [-\pi, \pi]^2 \mapsto [0, \pi]$ is defined as 
\begin{equation}
    d(\psi_1, \psi_2) = \min{(|\psi_1 - \psi_2|, 2\pi - |\psi_1 -\psi_2|)},
\end{equation}
to compute the difference between two angles.

\begin{figure}[t]
\centering
\includegraphics[width=0.48\textwidth]{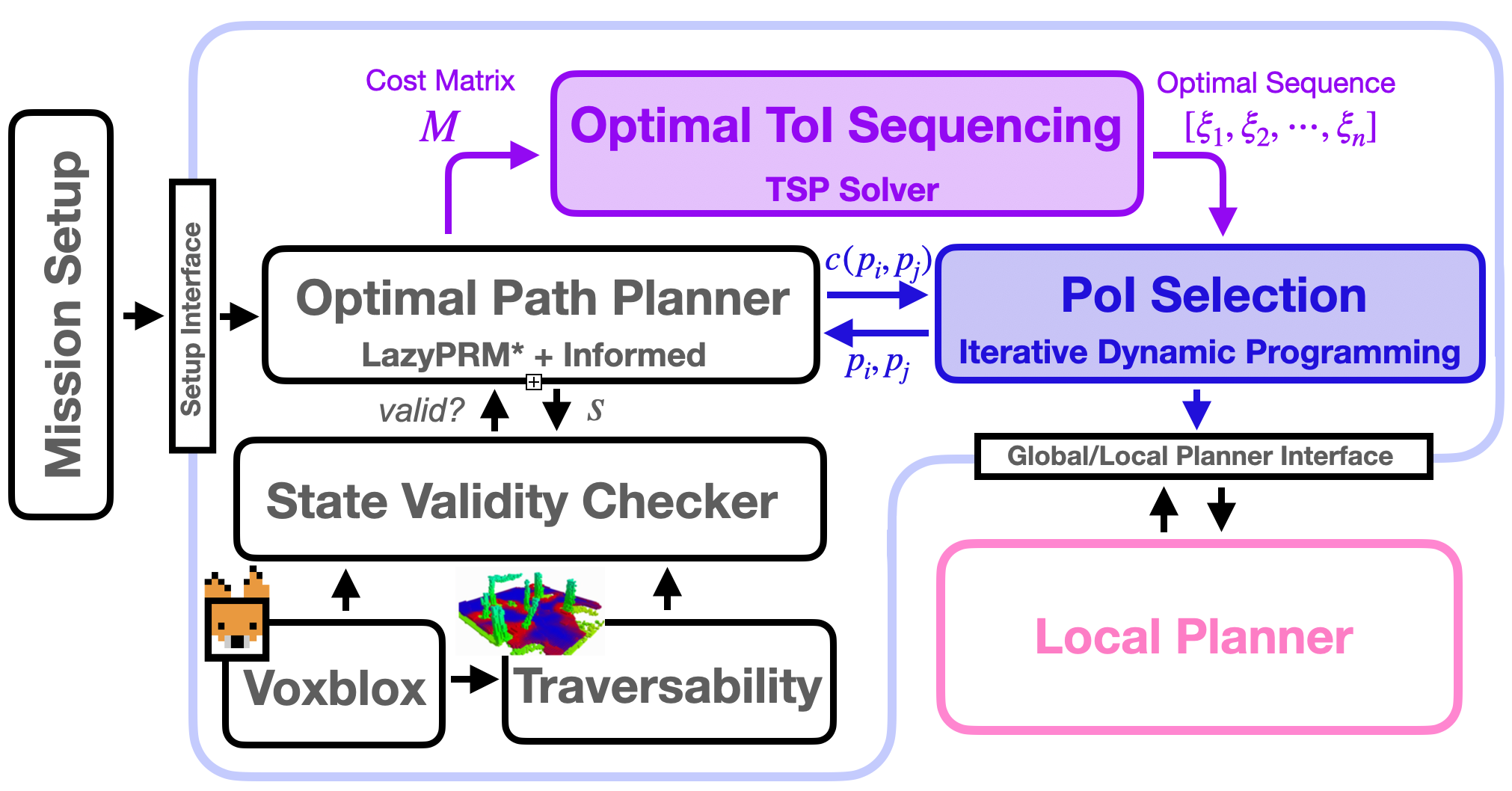}
\caption{System overview of the SMUG planner illustrating the information flow from accepting the mission to interacting with a low-level local planner. First, the optimal Target of Interest (ToI) sequence is computed~(purple, \cref{sec:tsp}). Second, the best Pose of Interest (PoI) is selected for each ToI~(blue, \cref{sec:idp}). Our sampling-based optimal path planner~(\cref{sec:opp}) is used in both steps to generate optimal collision-free paths. State validity is determined via geometric traversability checking in a volumetric map provided prior to the planning~(\cref{sec:validity}).}
\label{fig:overview}
\vspace{-0.6cm}

\end{figure}

\section{METHOD}
Solving the \gls{mtp} directly is challenging due to the large number of possibly collision-free paths. Therefore, we adopt a two-step method similar to~\cite{automatic}. In the first step, we determine the optimal sequence that visits every \gls{toi} by simplifying the problem to a \gls{tsp}. This simplification is valid, assuming that the \gls{poi}s are close to the respective \gls{toi}s. In the second step, we select the \gls{poi} at each \gls{toi} to minimize the total path cost. \cref{fig:overview} shows an overview of our system.
The system consists of the following modules: 
\begin{enumerate}
    \item An optimal path planner generating the optimal path connecting two states (\cref{sec:opp});
    \item A state validity checker using the \gls{tsdf} and the traversability map provided by Voxblox~\cite{voxblox} and a traversability module~\cite{lvn}, respectively, prior to the planning(\cref{sec:validity});
    \item A \gls{tsp} solver computing the optimal visiting sequence of the \gls{toi}s given a cost matrix containing the path cost between each pair of \gls{toi}s (\cref{sec:tsp});
    \item An \gls{idp} module selecting the optimal \gls{poi} for each \gls{toi} that minimizes the total path cost (\cref{sec:idp});
    \item A local planner~\cite{art} generating a finer path locally that follows the received global path.
\end{enumerate}

\subsection{Optimal Path Planning}
\label{sec:opp}
We use LazyPRM*~\cite{lazyprm*} with an informed sampler to generate the optimal collision-free paths between two states.

The sampler initially samples uniformly in $SE(2)$. After an initial path is found, it restricts the sample space based on the cost function and the current path cost to accelerate path optimization. Based on the informed sampler for the $L_2$ cost~\cite{informed}, we implement an informed sampler for the cost function $c(\cdot,\cdot)$~(\cref{eq:cost}).

Assume the path to find starts at state $s_1$ and ends at state $s_2$, and an initial path with cost $c_{0}$ is already found. The new sample $s_{new}$ are sampled in the ellipsoid
\begin{equation}
    \mathcal{E}\! = \!\left\{\!
    \begin{bmatrix}
        x \\ y
    \end{bmatrix}\!\Bigg|\!
    \left\|
    \begin{bmatrix}
        x \\ y
    \end{bmatrix}\! -\!
    \begin{bmatrix}
        x_{s_1} \\ y_{s_1}
    \end{bmatrix}\right\|^2_2
    \hspace{-1mm}\!+\!
    \left\|
    \begin{bmatrix}
        x \\ y
    \end{bmatrix}\! -\!
    \begin{bmatrix}
        x_{s_2} \\ y_{s_2}
    \end{bmatrix}\right\|^2_2\!\hspace{-1mm} <\! c_{0}\! -\! c_r\!
    \right\},
\end{equation}
where $c_r = d(\psi_{s_1}, \psi_{s_2})$.
Any sample $s$ outside this ellipsoid cannot improve the cost due to the following equation
\begin{equation}
\begin{aligned}
    &c(s_1,s) + c(s,s_2) \\
    = &\left\|
    \begin{bmatrix}
        x_{s_1} \\ y_{s_1}
    \end{bmatrix} -
    \begin{bmatrix}
        x_{s} \\ y_{s}
    \end{bmatrix}\right\|^2_2
    +\left\|
    \begin{bmatrix}
        x_{s} \\ y_{s}
    \end{bmatrix} -
    \begin{bmatrix}
        x_{s_2} \\ y_{s_2}
    \end{bmatrix}\right\|^2_2\\
    &+ d(\psi_{s_1}, \psi_{s}) + d(\psi_{s}, \psi_{s_2})\\
    \geq &\left\|
    \begin{bmatrix}
        x_{s_1} \\ y_{s_1}
    \end{bmatrix} -
    \begin{bmatrix}
        x_{s} \\ y_{s}
    \end{bmatrix}\right\|^2_2
    +\left\|
    \begin{bmatrix}
        x_{s} \\ y_{s}
    \end{bmatrix} -
    \begin{bmatrix}
        x_{s_2} \\ y_{s_2}
    \end{bmatrix}\right\|^2_2 + d(\psi_{s_1}, \psi_{s_2})\\
    \geq &c_0 - c_r +c_r
    = c_0.
\end{aligned}
\end{equation}
Then, the yaw angle $\psi_{s_{new}}$ is sampled uniformly in $[-\pi, \pi]$. This procedure is summarized below:
\RestyleAlgo{ruled}
\SetKwComment{Comment}{/* }{ */}
\begin{algorithm}[hbt!]
\footnotesize
\caption{\textit{Sample}$()$}
\label{alg:sample}
\eIf{path found}
{
$c_r \gets d(\psi_{s_1},\psi_{s_2})$\;
$c_0 \gets \textit{best cost}$\;
$x, y \gets \textit{SampleInformedL2}(s_1,s_2,c_0-c_r)$\;
}
{
$x \gets \textit{SampleUniform}(x_{min}, x_{max})$\;
$y \gets \textit{SampleUniform}(y_{min}, y_{max})$\;
}
$\psi \gets \textit{SampleUniform}(-\pi,\pi)$\;
$s_{new} \gets [x, y, \psi]^\top$\;
\Return{$s_{new}$}
\end{algorithm}

The sampled states are passed to a state validity checker to avoid states in collision with the environment.

\subsection{State Validity Checker}
\label{sec:validity}
We use a hierarchical obstacle avoidance scheme to ensure the feasibility and safety of the generated path efficiently. We adopt a robot-specific traversability module learned in simulation to accept the clearly safe states and discard clearly unsafe ones. Only states with unclear safety are checked for collision in an iterative volumetric fashion. 

\subsubsection{Traversability Filtering}
\begin{figure}[t]
\centering
\begin{subfigure}{.24\textwidth}
  \centering
  \includegraphics[width=0.9\textwidth]{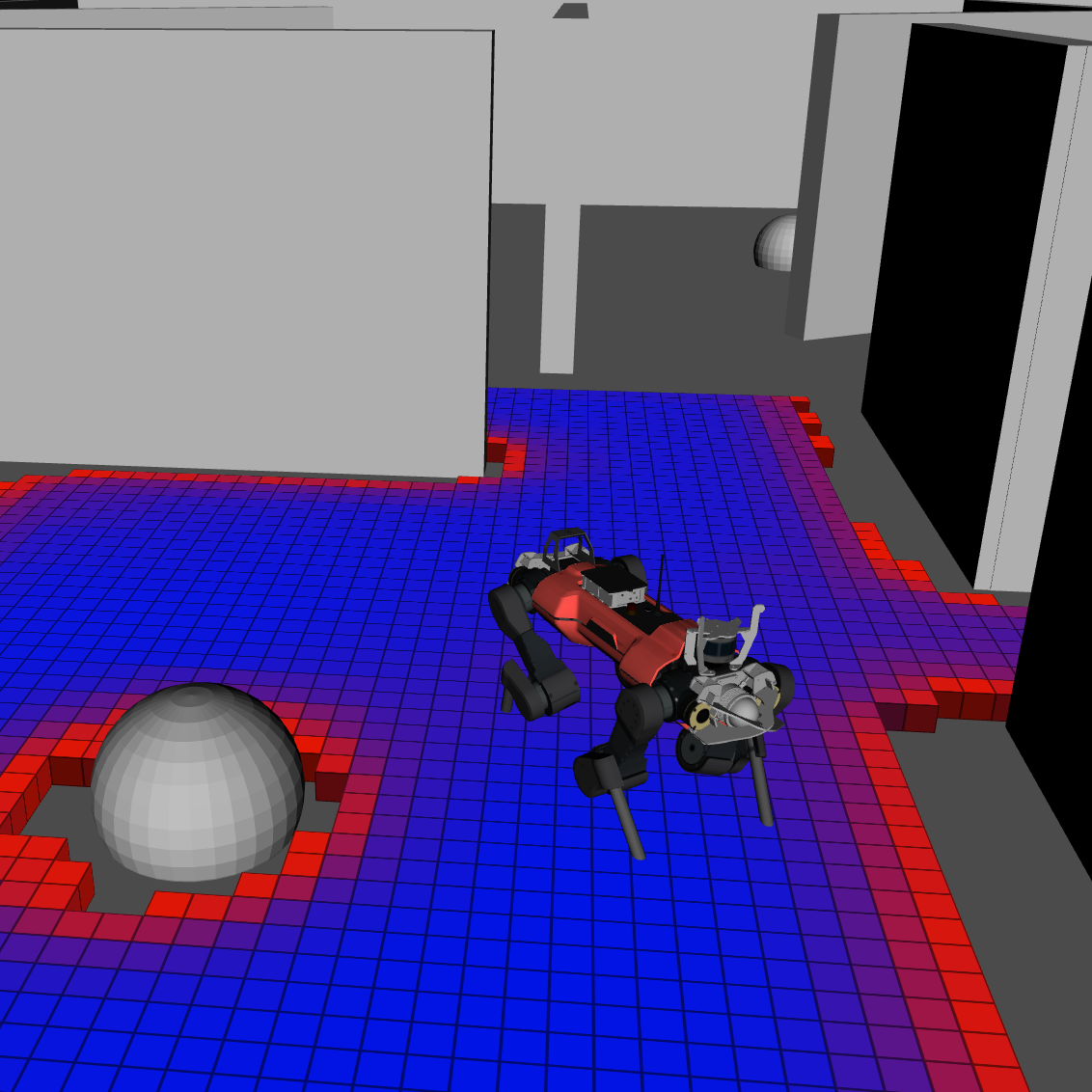}
    \caption{Traversability estimation}
    \label{fig:lvn}
\end{subfigure}%
\begin{subfigure}{.24\textwidth}
  \centering
  \includegraphics[width=0.9\textwidth]{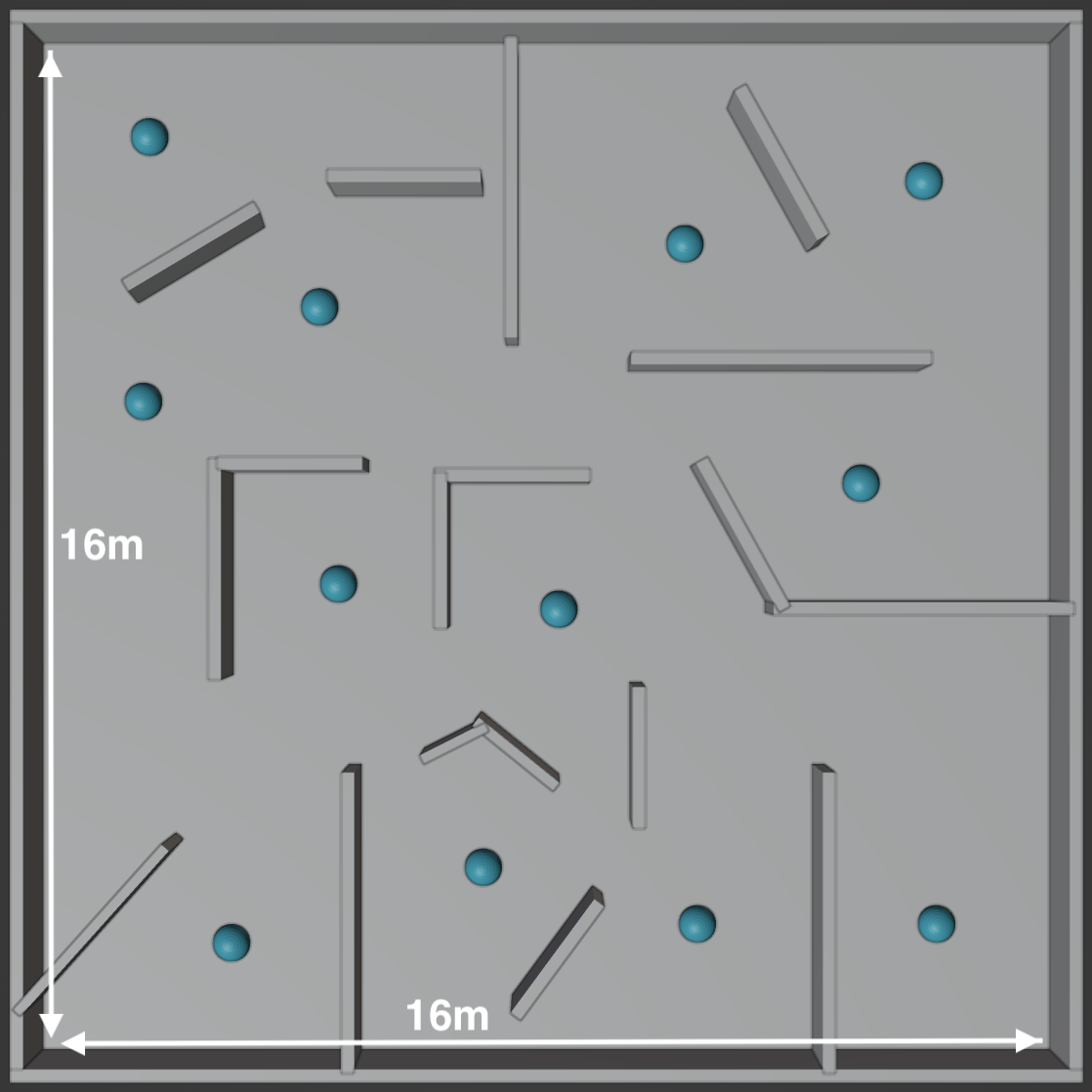}
    \caption{Sample environment}
\end{subfigure}
\vspace{0.3cm}

\begin{subfigure}{.24\textwidth}
  \centering
  \includegraphics[width=0.9\textwidth]{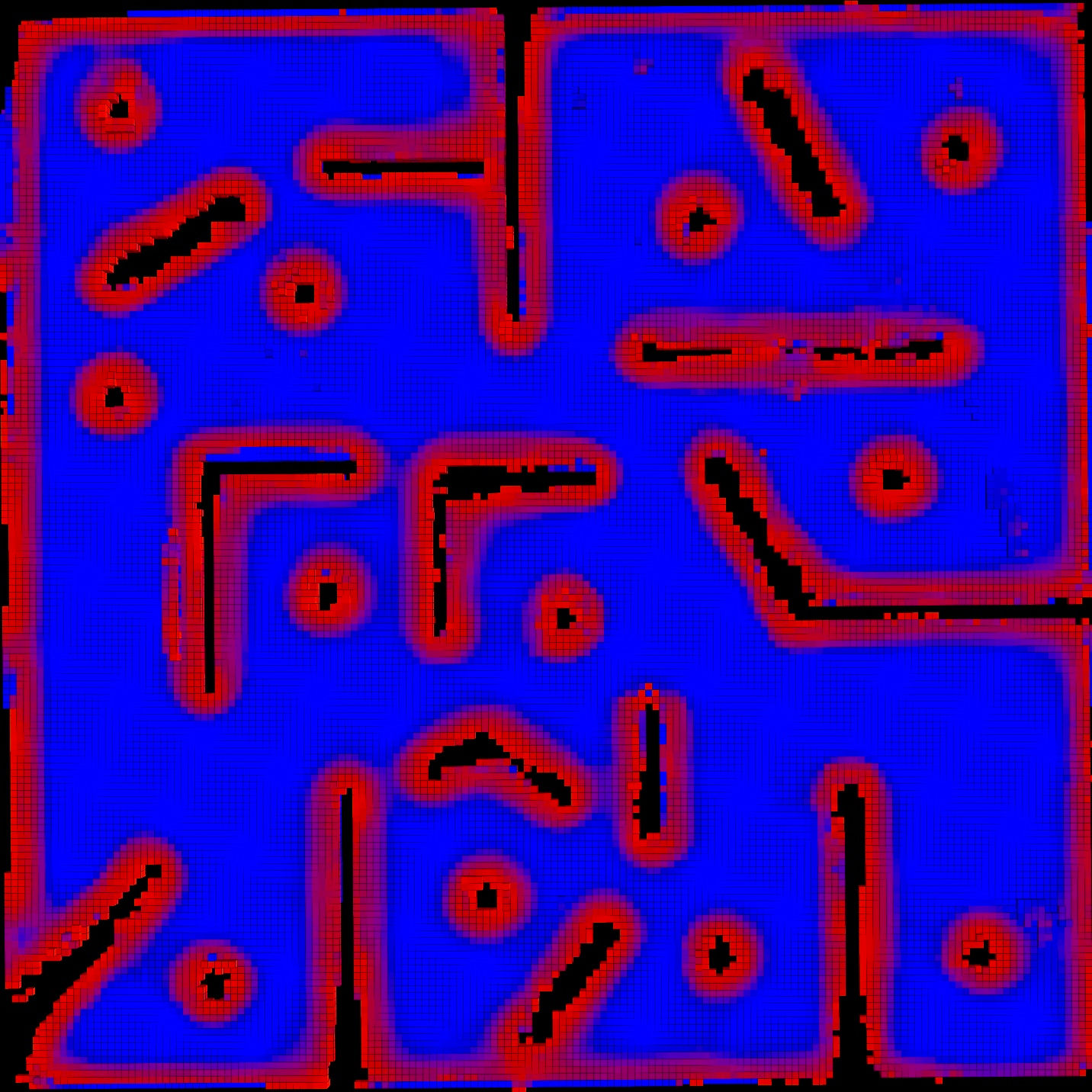}
    \caption{Traversability map}
\end{subfigure}%
\begin{subfigure}{.24\textwidth}
  \centering
  \includegraphics[width=0.9\textwidth]{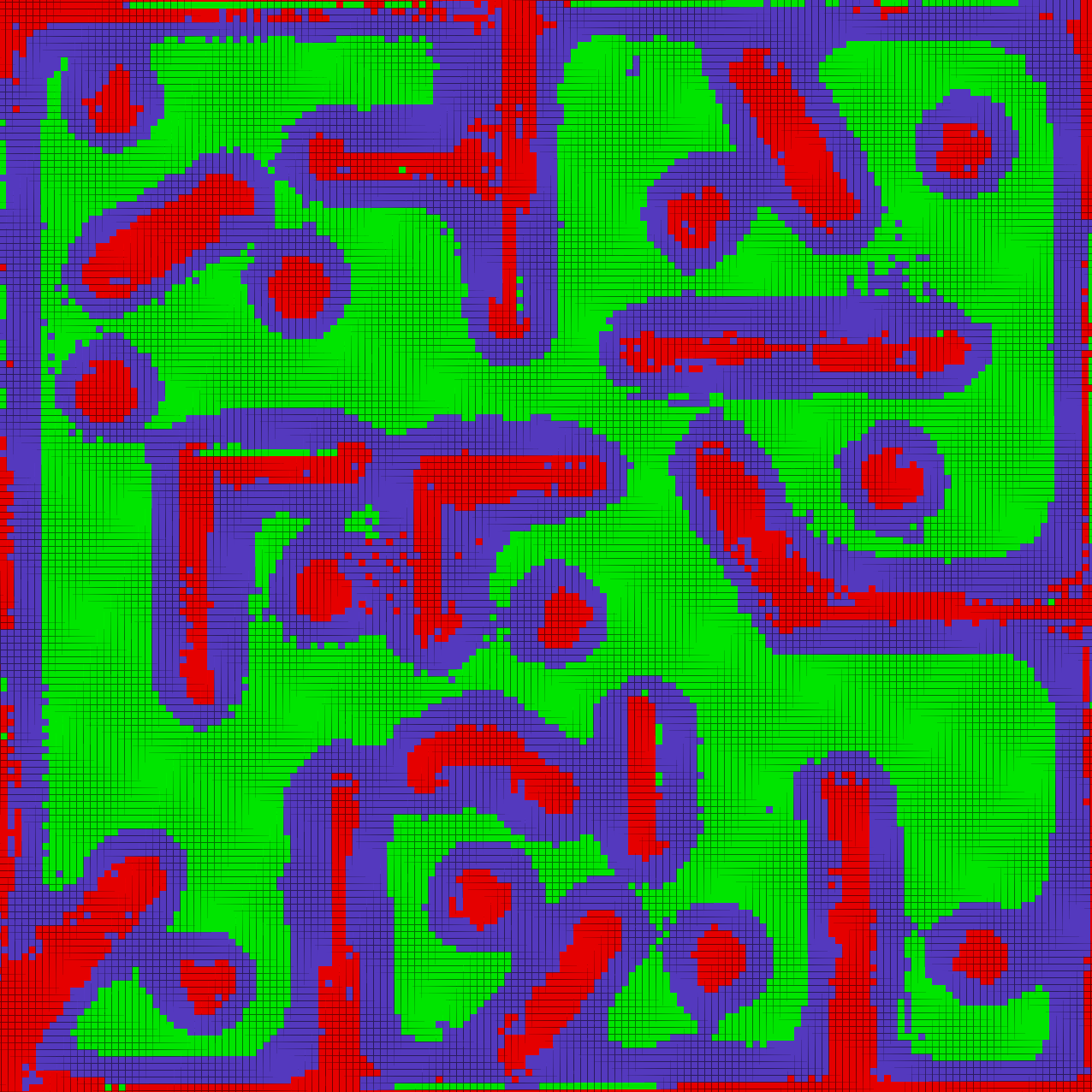}
    \caption{Traversability levels segmentation}
\end{subfigure}
\caption{(a) The traversability network does not only estimate the traversability for voxels, but it also segments the obstacles out. No traversability voxels are placed around the obstacles, which prevents the planner from sampling here. (b) A sample indoor environment with twelve targets. (c) Traversability map of the sample environment produced with the traversability network, the traversability ranges from $0$ to $1$, colored from red to blue. (d) Segmentation of the traversability map into three levels using $t_{low} = 0.3, t_{high} = 0.8$. High traversability is marked as green, low traversability as red, and intermediate traversability as purple.}
\label{fig:traversability}
\vspace{-0.6cm}

\end{figure}

We use a learned traversability map output by a convolutional neural network~\cite{lvn}, which takes the occupancy map as its input. The network assigns for each voxel a traversability estimate ranging from $0$ to $1$ based on the success rate of traversing through this voxel in simulation as illustrated in~\cref{fig:lvn}. We use this module to reduce the amount of iterative collision checking. The traversability is divided into three levels by setting two thresholds $t_{low}$ and $t_{high}$, satisfying $0\leq t_{low}\leq t_{high}\leq 1$. We obtained the traversability of a continuous state $s$ defined in \cref{eq:s} by querying the corresponding traversability voxel. States with traversability less than $t_{low}$ are directly invalidated, while those with traversability higher than $t_{high}$ are validated. Only states with traversability between $t_{low}$ and $t_{high}$ are checked for collisions. This avoids checking collisions for obviously valid states, for instance, the states in the middle of a wide open space, as well as definitely invalid states that lie within an obstacle.



\subsubsection{Volumetric Iterative Collision Checking}
\begin{figure}[t]
\centering
\begin{subfigure}{.12\textwidth}
  \centering
  \includegraphics[width=1\textwidth]{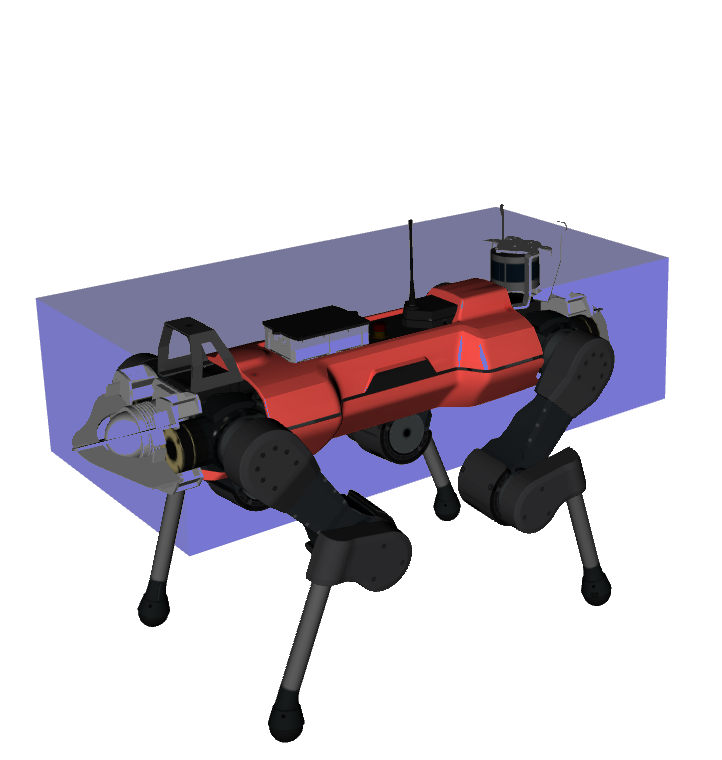}
    \caption{Bounding box}
\end{subfigure}%
\begin{subfigure}{.12\textwidth}
  \centering
  \includegraphics[width=1\textwidth]{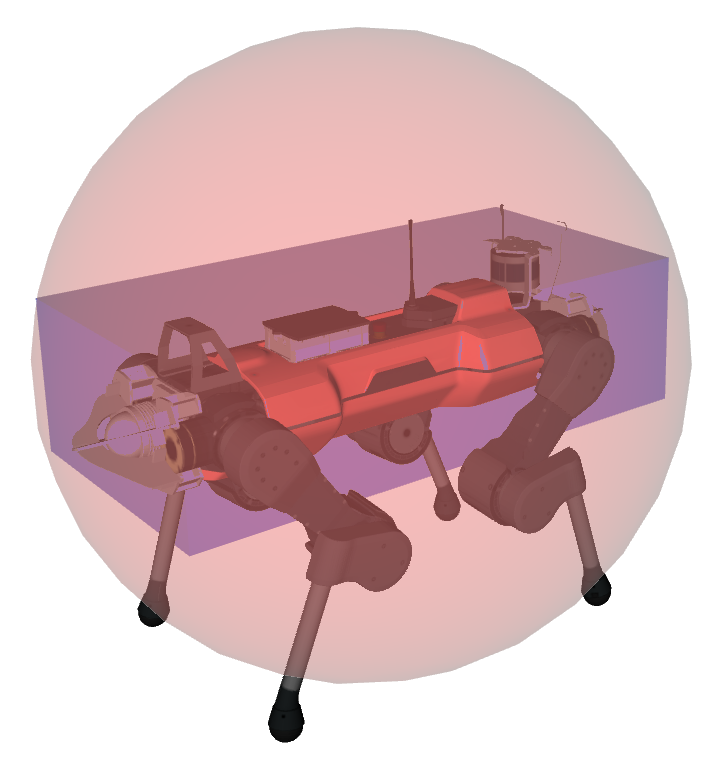}
    \caption{$1$st iteration}
\end{subfigure}%
\begin{subfigure}{.12\textwidth}
  \centering
  \includegraphics[width=1\textwidth]{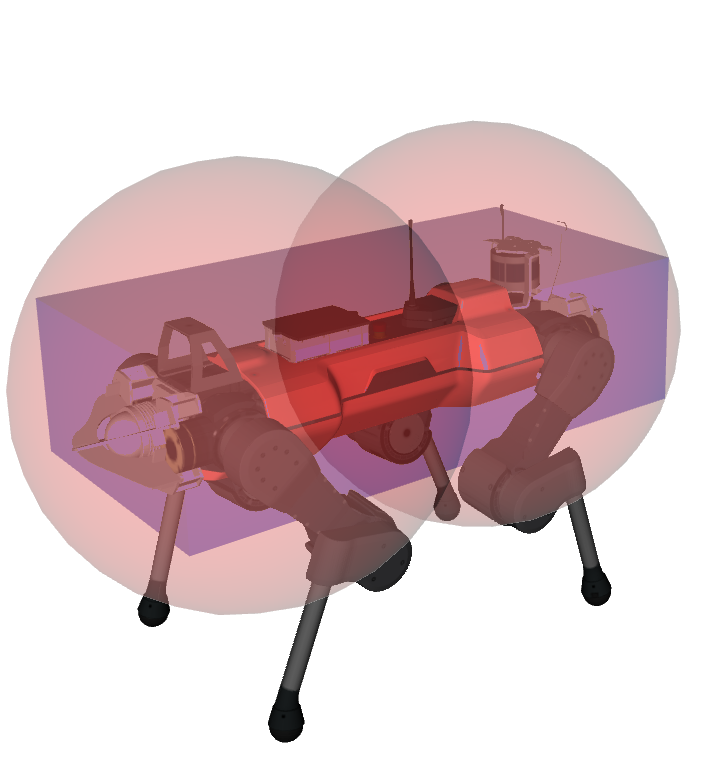}
    \caption{$2$nd iteration}
    \label{fig:collision2}
\end{subfigure}%
\begin{subfigure}{.12\textwidth}
  \centering
  \includegraphics[width=1\textwidth]{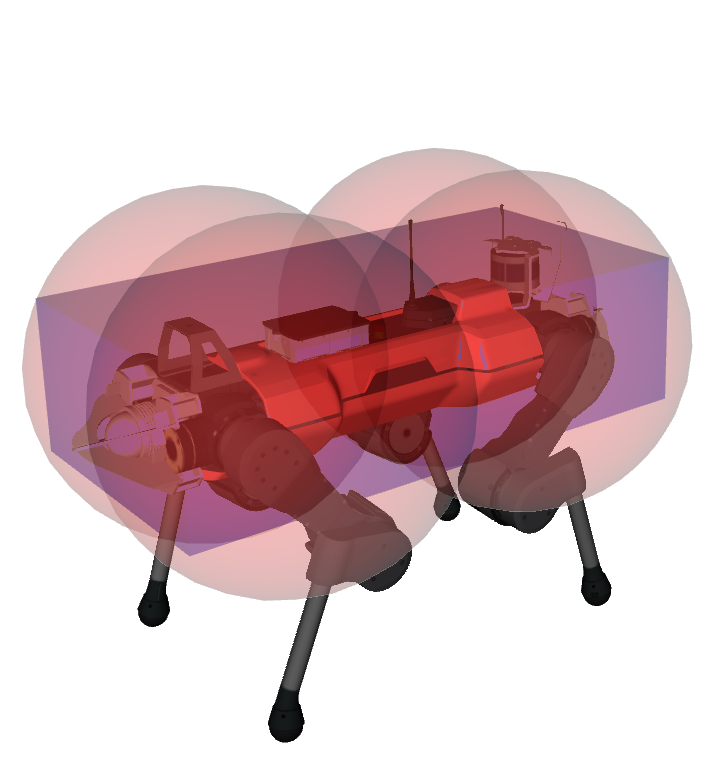}
    \caption{$3$rd iteration}
\end{subfigure}
\caption{The outer spheres (red) and the bounding box (blue) of the base of the ANYmal robot used in the first three iterations of the volumetric collision checking.}
\label{fig:collision}
\vspace{-0.6cm}
\end{figure}

For the region with unclear traversability, we resort to the iterative volumetric checking for the bounding box of the robot base. The method uses the \gls{tsdf} produced by Voxblox~\cite{voxblox} to check if the box is in collision with the environment. To do so, the distance from the box center to the nearest environment surface is obtained from the TSDF map and compared to the box's outer and inner sphere radius. The box is in collision if the distance is smaller than the inner sphere radius, and not in collision if the distance is larger than the outer sphere radius. Otherwise, no conclusion is reached, and the algorithm divides the current box into two sub-boxes along its longest dimension and performs the same procedure on them until a predefined minimal sub-box resolution is reached and only the outer sphere is evaluated for collision~(\cref{fig:collision}).


\subsection{Optimal Sequencing (\gls{tsp})}
\label{sec:tsp}
To find the optimal visiting sequence of the \gls{toi}s, we first plan the collision-free paths connecting every pair of \gls{toi}s, resulting in the following cost matrix:
\begin{equation}
    M = 
    \begin{bmatrix}
        &0  &m_{0,1} &m_{0,2} &\cdots &m_{0,n}\\
        &m_{1, 0} &0 &m_{1,2} &\cdots &m_{1,n} \\
        &\vdots & &\ddots & &\vdots \\
        &m_{n,0} &m_{n,1} &m_{n,2} &\cdots &0
    \end{bmatrix},   
\end{equation}
where $m_{i,j}$ is the cost of the collision-free path from $t_i$ to $t_j$. For simplicity, we use $t_0 = s$. This matrix is passed to a TSP solver~\cite{google} computing a near-optimal visiting sequence attempting to minimize:
\begin{equation}
    \sum_{i=0}^{n-1}m_{\xi_{i},\xi_{i+1}} + m_{\xi_{n},\xi_{0}}
\end{equation}
with $\xi_0 = 0$ and $[\xi_1, \cdots, \xi_n]^\top$ being a permutation of $[1,\cdots,n]$.

From now on we denote the $i$-th visited \gls{toi} as $t_i$, where $t_0 = s$. For each $t_i$, we denote the corresponding set of \gls{poi}s as $\mathcal{P}_i$. The collision-free path connecting two poses $p_i, p_j$ is referred to as $W(p_i, p_j)$ in the following.

\subsection{Iterative Dynamic Programming}
\label{sec:idp}
After the visiting sequence is determined, the planner selects the \gls{poi} for each \gls{toi}. Given all collision-free paths between every \gls{poi} of two consecutively visited \gls{toi}, i.e. $W(p_{i,j},p_{i+1,k}), \forall{p_{i,j}}\in \mathcal{P}_i, {p_{i+1,k}}\in\mathcal{P}_{i+1}$, the optimal path cost can be found via \gls{dp} by solving the following equations

\begin{align}
\begin{split}
    J_n^*(p_{n,j}) ={}& Cost(W(p_{n,j}, s)),
    \label{eq:dp1}
\end{split}\\
\begin{split}
    J_i^*(p_{i,j}) ={}& \min_{p\in\mathcal{P}_{i+1}}Cost(W(p_{i, j},p)) + J_{i+1}^*(p),\\
    &\hspace{100pt} \forall i\in\left\{1,\cdots, n-1\right\},
    \label{eq:dp2}
\end{split}\\
\begin{split}
    J_0^*(s) ={}& \min_{p\in\mathcal{P}_{1}}{Cost(W(s, p)) + J_{1}^*(p)}.
    \label{eq:dp3}
\end{split}
\end{align}
The optimal \gls{poi}s to choose are thus obtained with
\begin{align}
\begin{split}
    \hspace{25pt}p_0^*  ={}& s,
    \label{eq:dp4}
\end{split}\\
\begin{split}
p_{i}^* ={}& \argmin_{p\in\mathcal{P}_{i}}Cost(W(p^*_{i-1},p)) + J_{i}^*(p),\\
&\hspace{100pt}\forall i\in\left\{1,\cdots, n\right\}.
    \label{eq:dp5}
\end{split}
\end{align}

\gls{dp} is guaranteed to output the global optimum given all paths between \gls{poi}s of two consecutively visited \gls{toi}s. However, in the case of $N$ \gls{toi}s and $M$ \gls{poi}s per \gls{toi}, i.e. $|\mathcal{T}| = N, |\mathcal{P}_i| = M$, it requires planning $M^2(N-1)+2M$ collision-free paths which increases quadratically with the number of \gls{poi}s. 
Eventually, not every path is used to form the optimal global path. 
Therefore, we propose to use \gls{dp} in an iterative fashion and only plan the collision-free paths if they are needed to construct the optimal global path after each iteration.
We define:
\begin{equation}
    l(p_i, p_j) = 
    \begin{cases}
        Cost(W(p_i, p_j)), &\mbox{if collision-free path } \\
        &W(p_i, p_j)\mbox{ is planned},\\
        c(p_i, p_j), &\mbox{otherwise}.
    \end{cases}
\end{equation}
This is the path cost of the collision-free $W(p_i, p_j)$ path between $p_i$ and $p_j$, if it is already planned. Otherwise, $l(p_i,p_j)$ approximates the actual cost of $W(p_i, p_j)$ using the straight line between $p_i$ and $p_j$, i.e. $c(p_i, p_j)$ defined in~\cref{eq:cost}, which is a lower bound of $Cost(W(p_i, p_j))$.

At each iteration, we perform the introduced \gls{dp}~(\cref{eq:dp1,eq:dp2,eq:dp3,eq:dp4,eq:dp5}), except replacing $Cost(W(\cdot, \cdot))$ by $l(\cdot, \cdot)$, to find the \gls{poi}s. Then we generate the collision-free paths connecting the two consecutively visited \gls{poi}s and update $l$ accordingly. 
The algorithm terminates if the selected \gls{poi}s do not change over iterations.
In this way, not necessarily all $M^2(N-1)+2M$ collision-free paths need to be generated, reducing the total planning time. \gls{idp} preserves the optimality of \gls{dp} given the same \gls{poi} to \gls{poi} paths because the resulting path cost is lower than the lower bound of any other possible path.

Finally, we concatenate the path segments according to the obtained sequence of \gls{toi} and the selected \gls{poi}. We align the heading of the waypoints with the path direction (except for the ones at \gls{poi}s, which are fixed) to ensure that the robot moves forwards instead of sideways.

\section{Experiments}
\subsection{Simulation}
\begin{figure}[h]
\centering
\begin{subfigure}{.24\textwidth}
  \centering
  \includegraphics[width=0.96\textwidth]{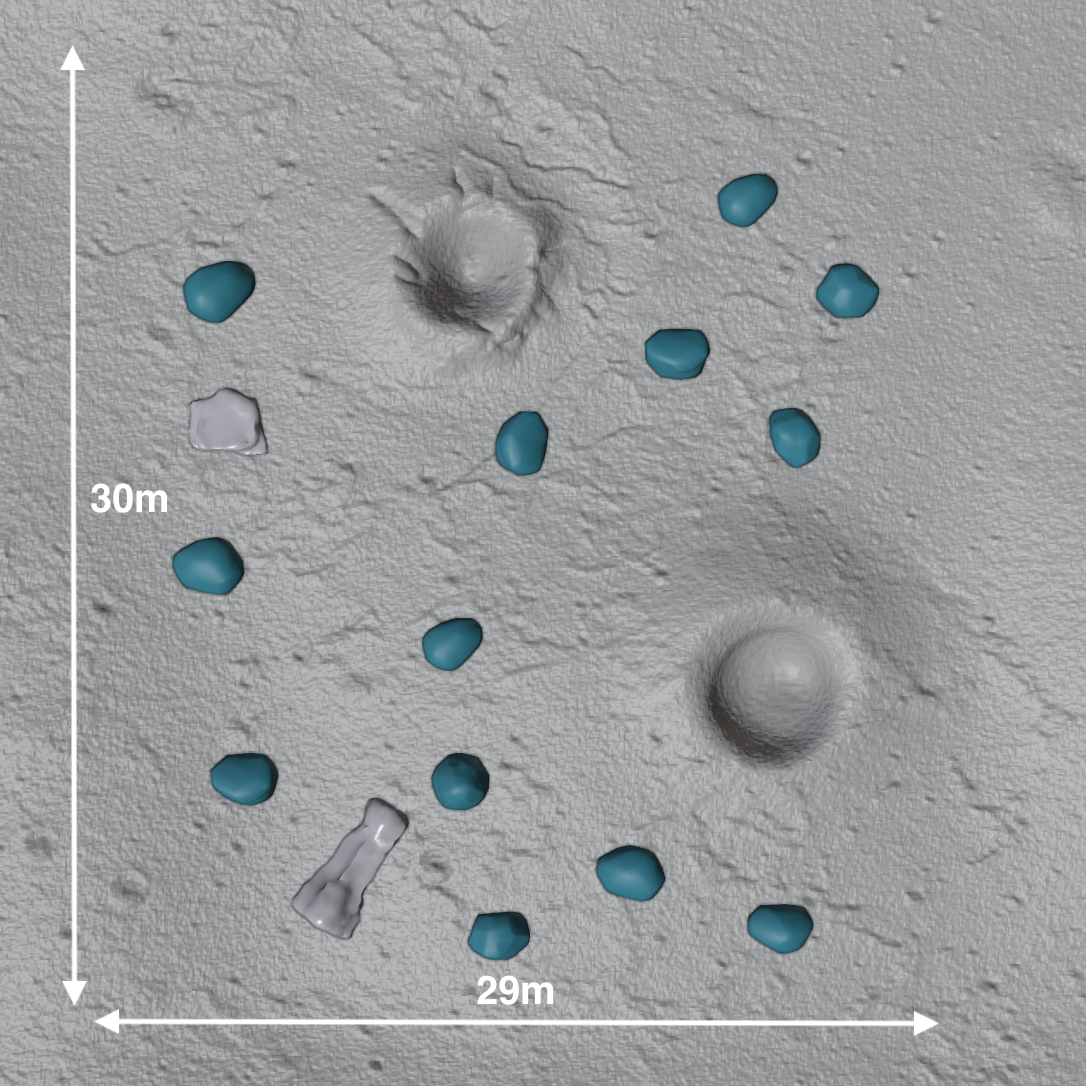}
    \caption{Lunar}
    \label{fig:env lunar}
\end{subfigure}%
\begin{subfigure}{.24\textwidth}
  \centering
  \includegraphics[width=0.96\textwidth]{figure/test_env2.png}
    \caption{Indoor}
    \label{fig:env indoor}
\end{subfigure}
\caption{(a) A lunar analog environment of $30m \times 29m$ with 13 stones to investigate (blue). (b) An indoor simulation environment of $16m \times 16m$ with 12 targets to inspect (blue)}
\label{fig:env}
\end{figure}

We test our planner in two customized simulation environments, a lunar analog environment~(\cref{fig:env lunar}) and an indoor space~(\cref{fig:env indoor}), to illustrate its applicability in both exploration and industrial settings.

\begin{figure}[t]
\centering
\begin{subfigure}{.24\textwidth}
  \centering
    \includegraphics[width=0.96\textwidth]{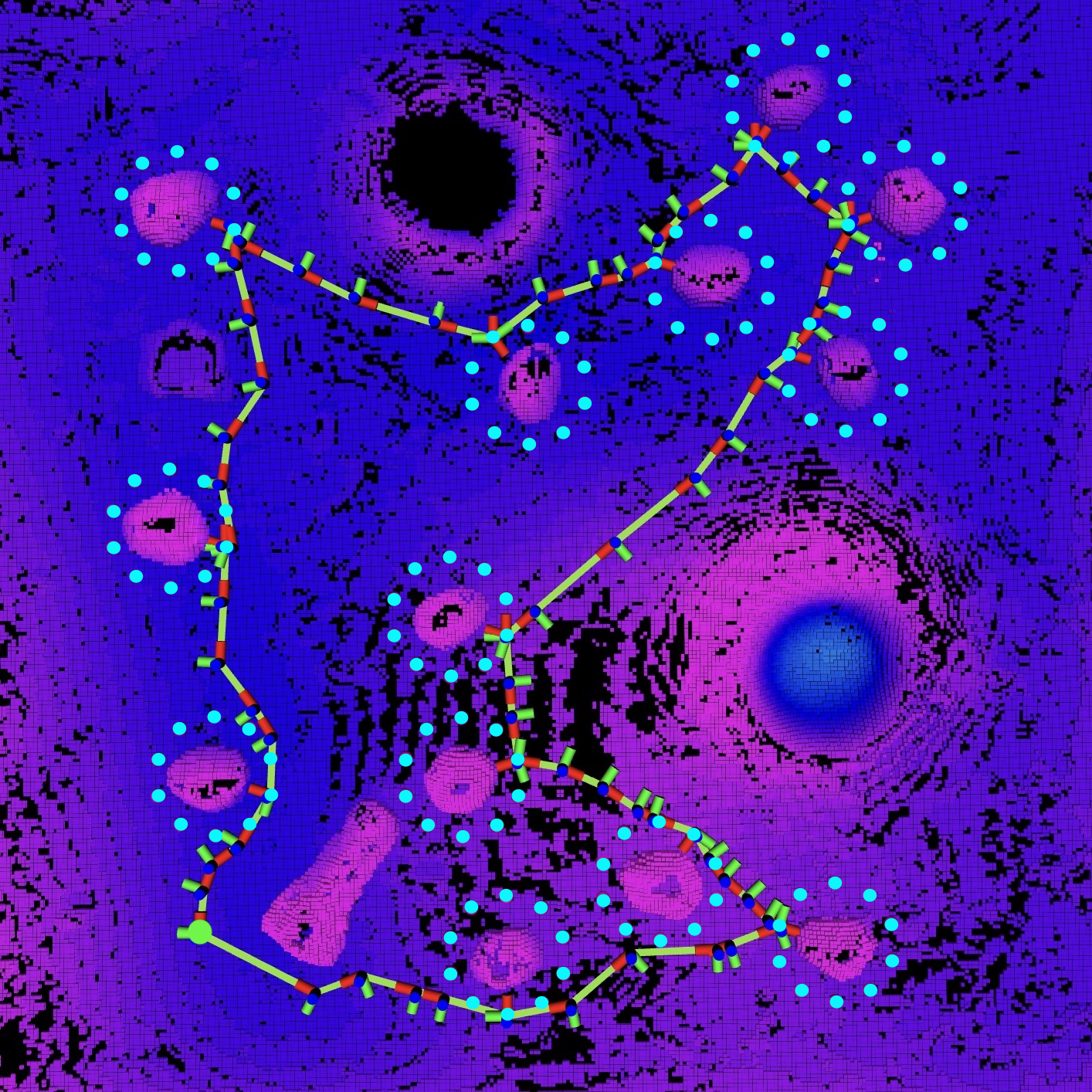}
\caption{Lunar}
\end{subfigure}%
\begin{subfigure}{.24\textwidth}
  \centering
  \includegraphics[width=0.96\textwidth]{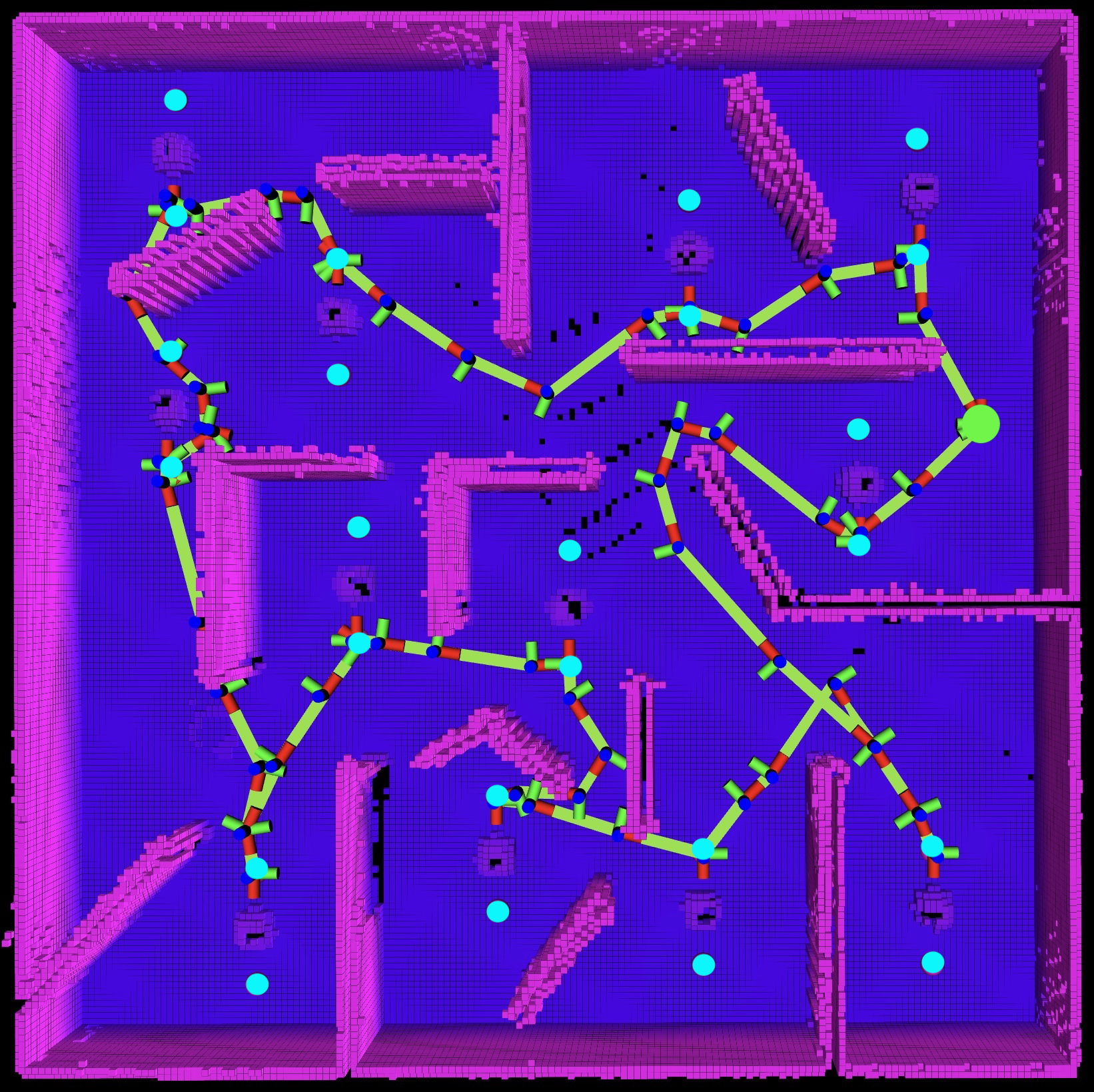}
\caption{Indoor}
\end{subfigure}
\caption{Samples of generated paths for the lunar and the indoor environments, respectively. The feasible poses to choose are marked as cyan dots, and the robot start poses are marked as green dots.}
\label{fig:path}
\vspace{-0.6cm}

\end{figure}

In the lunar environment, there are 13 \gls{toi}s, and ten \gls{poi}s are generated uniformly around each. In the indoor environment, we set 12 \gls{toi}s and for each two \gls{poi}s to choose from to represent smaller-size problems. \cref{fig:path} shows the generated path for the environments.
Since no other method exists that solves the presented problem, we compare the proposed method with the baseline designed ourselves. The following experiments are conducted on a laptop with Intel Core i7-9750H CPU and 32GB RAM. All means and standard deviations are calculated from ten consecutive runs.

\subsubsection{\gls{idp} / \gls{dp} / \gls{irba} Comparison}
We compare the performance of the \gls{idp} with the method of preplanning all $M^2(N-1)+2M$ collision-free paths before performing \gls{dp} once, which we refer to as \gls{dp}, and an adaption of the method proposed in~\cite{automatic} used to solve \gls{mtpgr}, which we refer to as Iterative RBA (IRBA).

Since \gls{idp} and \gls{irba} do not necessarily require planning all $M^2(N-1)+2M$ paths, the planning time is largely reduced, as shown in \cref{tab:idp}. In the lunar environment with $13$ \gls{toi}s and $10$ \gls{poi}s each, \gls{dp} needs to plan $1220$ collision-free paths in the second stage and spends on average $71.87$ seconds in total, whereas \gls{idp} only needs $8.05$ seconds on average, saving $88.80\%$ of the planning time and is $8.9$ times faster. This improvement is less noticeable in the indoor environment with $12$ \gls{toi}s and $2$ \gls{poi}s each, where \gls{dp} preplans $48$ collision-free paths spending in total an average of $7.59$ seconds and \gls{idp} uses $7.20$ seconds, reducing the planning time only by $5.1\%$ due to the small number of \gls{poi}. The \gls{irba} and \gls{idp} have roughly the same performance in terms of planning time. The differences are within $8\%$ and $3\%$ of standard deviation in the two environments. 

As for the path cost, \gls{idp} produces a similar cost as \gls{dp} in both tests (\cref{tab:idp}). The difference is only due to the denser graph that \gls{dp} constructs while planning more paths.
However, \gls{irba} increases the cost compared to \gls{dp} by $2-3\%$, and does not provide any theoretical optimality guarantee compared to \gls{idp}, therefore in different environments, the performance gap \gls{irba} and \gls{idp} may be larger. 
When running the experiment on the same graph, \gls{idp} produces the same path cost as \gls{dp} as explained in~\cref{sec:idp}, while \gls{irba} often produces worse results.

\begin{table}
\centering
\begin{tabular}{@{}ccccccccc@{}}\toprule
& & \multicolumn{3}{c}{Lunar} & \phantom{abc}& \multicolumn{3}{c}{Indoor}\\
\cmidrule{3-5} \cmidrule{7-9}
& & DP & IDP & IRBA && DP & IDP & IRBA\\\midrule
\parbox[t]{2mm}{\multirow{2}{*}{\rotatebox[origin=c]{90}{Time}}} & Mean & 71.87& \textbf{8.05} & 8.25 && 7.59& 7.20 &\textbf{7.18}\\
& Std. & 8.89 & 2.50 & 2.78 && 1.35 & 1.00  &0.59\\ \midrule
\parbox[t]{2mm}{\multirow{2}{*}{\rotatebox[origin=c]{90}{Cost}}} & Mean & \textbf{97.22}& 97.58 &100.08 && \textbf{82.05} & 82.85 & 83.66\\
& Std. & 1.16 & 1.83 &2.33 && 3.75 & 3.70 &4.02  \\
\bottomrule
\end{tabular}
\caption{Comparison of \gls{dp} (brute force optimal path), \gls{idp} (ours) and \gls{irba} (adapted from~\cite{automatic}). \gls{idp} generates slightly higher cost than \gls{dp} because \gls{dp} has a denser graph while planning all paths.}
\label{tab:idp}
\end{table}

\subsubsection{Hierarchical Obstacle Avoidance}

\begin{table}
\centering
\begin{tabular}{@{}cccccc@{}}\toprule
&& \multicolumn{1}{c}{$t_{high} = 0.3$} & \multicolumn{1}{c}{$t_{high} = 0.8$} & \multicolumn{1}{c}{Full Collision} \\
&& \multicolumn{1}{c}{$t_{low} = 0.3$} & \multicolumn{1}{c}{$t_{low} = 0.3$} & \multicolumn{1}{c}{Checking} \\ \midrule
\parbox[t]{2mm}{\multirow{2}{*}{\rotatebox[origin=c]{90}{Time}}} & Mean & \textbf{8.05} & 9.39 & 11.56\\
& Std. & 2.50 & 3.61 & 2.29\\ \midrule
\parbox[t]{2mm}{\multirow{2}{*}{\rotatebox[origin=c]{90}{Cost}}} & Mean & 97.58 &97.87 & \textbf{97.17}\\
& Std. & 1.83 & 1.90 &1.79\\
\bottomrule
\end{tabular}
\caption{Comparision between traversability threshold and full collision checking in the lunar environment.}
\label{tab:trav}
\vspace{-0.6cm}

\end{table}

\begin{table}
\centering
\begin{tabular}{@{}lccc@{}}\toprule
 & \multicolumn{2}{c}{$t_{high} = t_{low} = 0.3$} & \multicolumn{1}{c}{Full Collision Checking} \\ \midrule
A* & \multicolumn{2}{c}{5.05} & 4.79 \\
Sampling & \multicolumn{2}{c}{0.43} & 0.84 \\
\textbf{Edge checking} & \multicolumn{2}{c}{\textbf{2.32}} & \textbf{5.68} \\
Vertex checking & \multicolumn{2}{c}{0.01} & 0.05 \\
Miscellaneous & \multicolumn{2}{c}{0.24} & 0.23 \\ \midrule
Total & \multicolumn{2}{c}{8.05} & 11.56 \\
\bottomrule
\end{tabular}
\caption{Breakdown of the required time of each computation step.}
\label{tab:break}
\end{table}

\begin{figure*}[t!]
    \centering
    \includegraphics[width=1\textwidth]{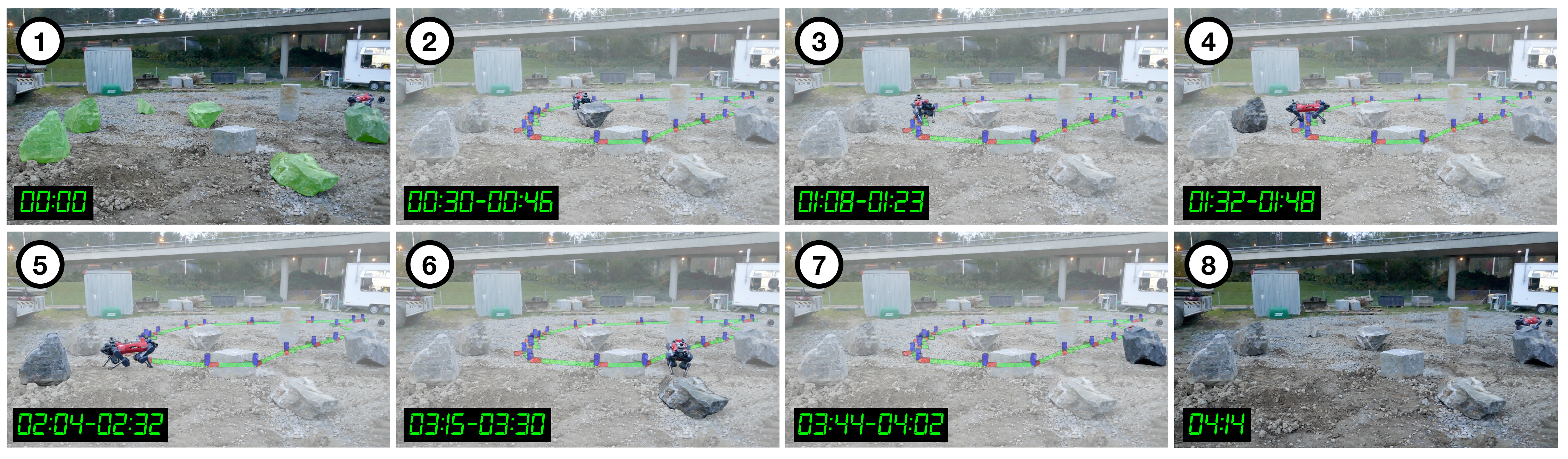}
    \caption{Real-world deployment on the ANYmal quadruped robot. The subfigures show the snapshots of the entire mission. The six specified targets are marked in green in (1). The robot follows the global path to perform inspection near each target one by one (2-7, with the current robot inspection pose and target highlighted in a dimmed background). The time span where the robot stays at each target is given in each picture relative to the mission start, each of which includes 15 seconds for inspection and the computation time of the local planner generating the next path. Finally, the robot homes in (8).}
    \label{fig:exp}
\vspace{-0.3cm}

\end{figure*}

\begin{figure}[t!]
\centering
    \includegraphics[width=0.48\textwidth]{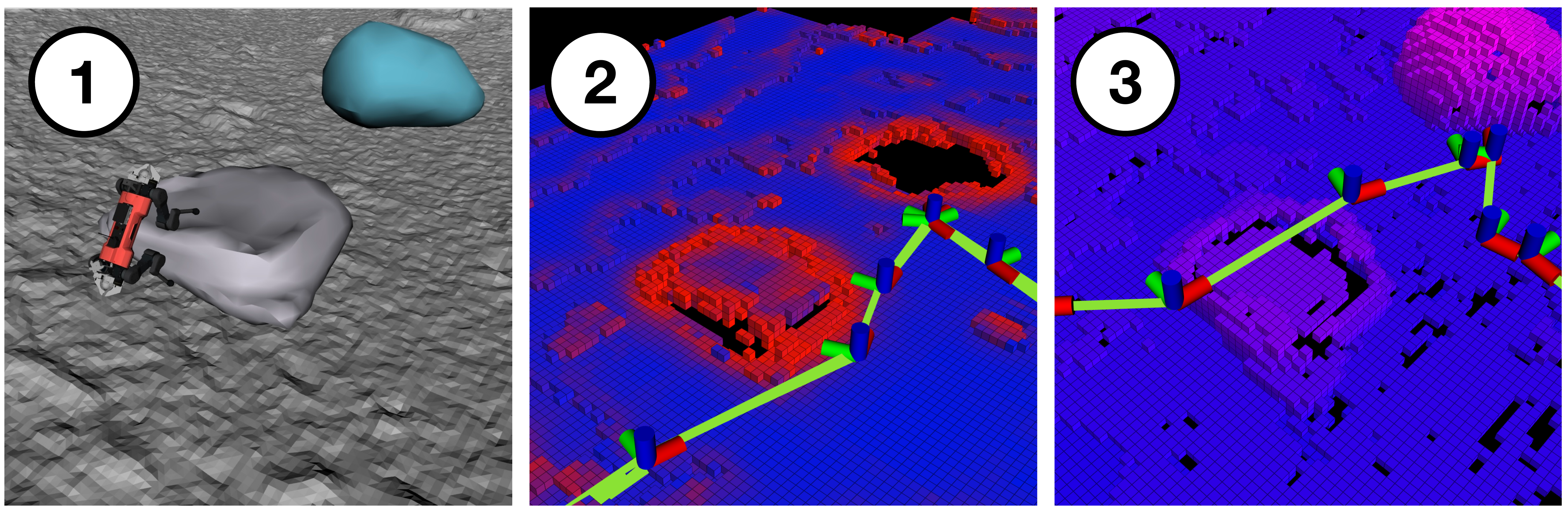}
    \caption{(1) Risky region for the robot to traverse. (2) The path planned with the traversability filtering avoids the risky zone. (3) The path planned with only collision-checking choose to pass over it. Although there is no collision with the environment, the path is risky.}
    \label{fig:safety}
\vspace{-0.3cm}

\end{figure}

We evaluate the effectiveness of the hierarchical obstacle avoidance scheme in the lunar simulation environment~(\cref{fig:env lunar}). The environment contains two obstacles, 13 stones (\gls{toi}s), two craters and various slopes, resulting in various traversability and making it more challenging and realistic than the indoor scene. As shown in~\cref{tab:trav}, with the use of the traversability module, $30\%$ of the planning time is saved due to fewer voxel queries of the state validity checker. We set the collision checking for the robot base to stop at the second iteration, as there the outer spheres are already a sufficiently good approximation of the ANYmal robot's base (\cref{fig:collision2}). The acquisition of the traversability of a state only requires one voxel query, whereas iteratively checking collision for that state requires at most four queries (when stopping at the second iteration). 
Traversability filtering marginally increases the geometric path cost but leads to a safer path. This is illustrated in \cref{fig:safety};3, where the planned path with full collision-checking passes over the stone, given that the poses do not incur any collision. On the other hand, the traversability module can account for the risk and filters out the high-risk areas, avoiding passing over the stone.

The breakdown of the timings in~\cref{tab:break} shows that the improvement of the planning time mostly originates from edge checking, which requires a large amount of vertex checking and hence many more voxel queries if using volumetric collision checking.

\begin{figure}[t!]
    \centering
    \includegraphics[width=0.48\textwidth]{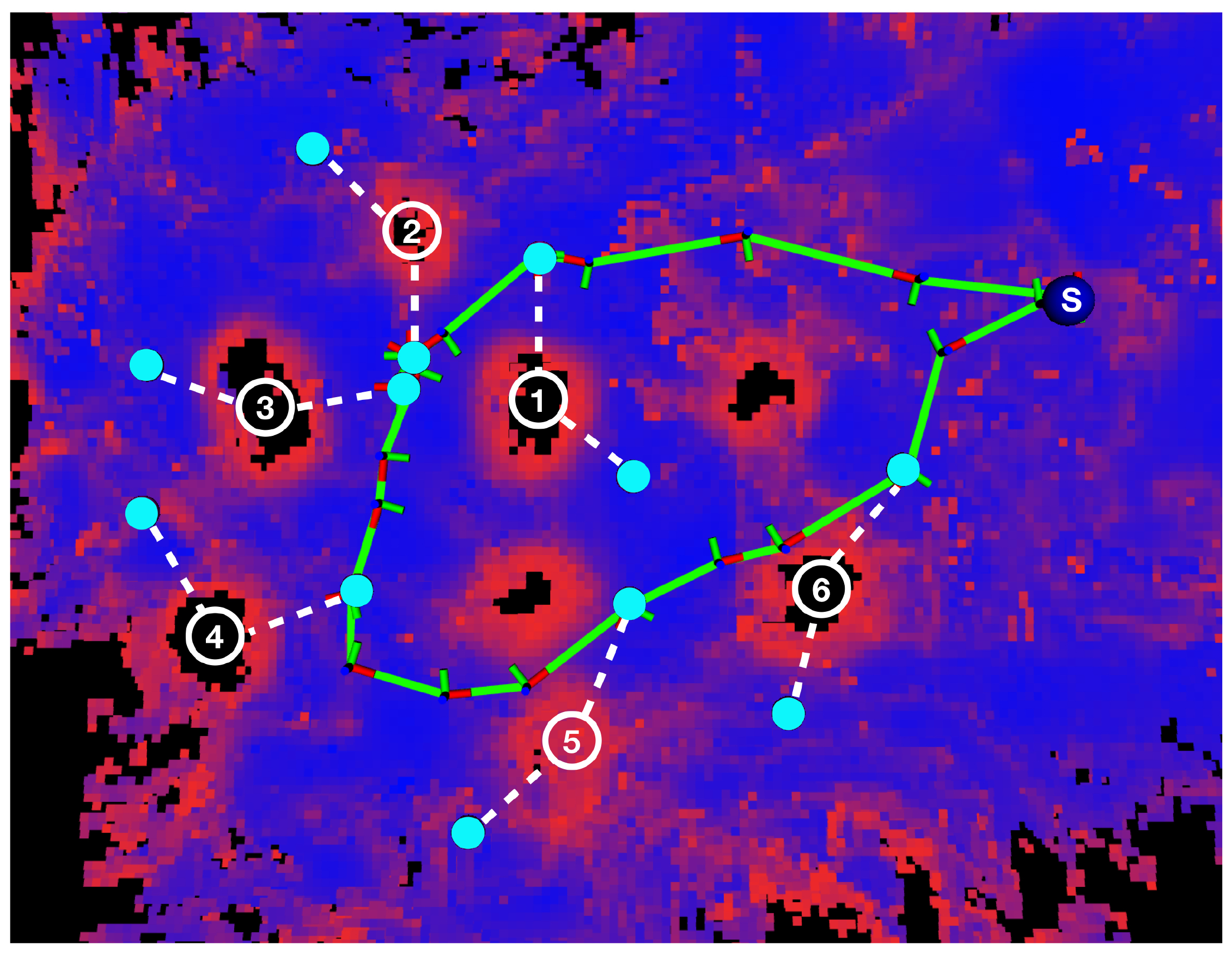}
    \caption{Generated path (green) for the real-world experiment. The \gls{toi}s are labeled with white numbers and connected to their respective \gls{poi}s (marked as cyan dots) with white dashed lines. The robot start pose is marked as a blue dot and labeled with $S$.}
    \label{fig:exp9}
    \vspace{-0.6cm}

\end{figure}

\subsubsection{Scalibility}
\begin{table}
\centering
\begin{tabular}{cccccccc}
\toprule
\parbox[t]{2mm}{\multirow{2}{*}{ToI}} & \multicolumn{3}{c}{Time} & \multicolumn{3}{c}{Cost}\\
\cmidrule(lr){2-4} \cmidrule(lr){5-7}
& DP & IDP & IRBA & DP & IDP & IRBA \\
\midrule
6  &5.42 & 1.05&\textbf{0.74} &\textbf{64.55} &65.17 &66.93\\
12 & 20.12 & \textbf{5.12} &5.82  &\textbf{87.75} &89.04  & 92.04 \\
24 & 80.49 & 22.31 & \textbf{19.67} &\textbf{122.55} & 123.27  & 124.10 \\
48 & 508.71 & \textbf{176.17} & 180.42  &164.04 &\textbf{163.15} & 167.55 \\
\bottomrule
\end{tabular}
\caption{The average computation time and path cost of DP, IDP and RBA when the number of ToI varies.}
\label{tab:toi}
\vspace{-0.6cm}

\end{table}

To show the scalability of the method, we deploy the proposed planner in the lunar environment to missions of different numbers of \gls{toi}s while keeping the \gls{poi} number of each \gls{toi} fixed at six. As shown in~\cref{tab:toi}, the planner is able to plan through 48 \gls{toi}s within three minutes, saving $60\%$ to $80\%$ of planning time compared to \gls{dp} while having similar path costs. The planning time of \gls{idp} increases roughly quadratically as the number of \gls{toi} increases from $6$ to $24$, and more than quadratically from $24$ to $48$. This is because the planning time does not only depend on the number of paths but also on the difficulty of individual paths and the size of the incrementally built planning graph.

\subsection{Real World Deployment}
We deploy the planner on the ANYmal quadruped robot in a real-world outdoor multi-goal mission on pitted terrain covered by gravel (\cref{fig:exp}). Six stones with irregular shapes are specified as \gls{toi} and two cuboid-shaped stones are placed on the terrain as obstacles. 

We first command the robot to walk in the environment to build the needed TSDF and traversability map. Then we select for each \gls{toi} two \gls{poi}s to choose from. The planner generates the optimal global path accordingly. Thereafter, the path segment consisting of $SE(3)$ waypoints from the current position to the next target is sent to a low-level local planner~\cite{art}, which generates the local path following the waypoints. We give $15$ seconds to investigate each \gls{toi}. Once the inspection at one \gls{toi} is finished, the global path to the next target is sent to the local planner. The robot is able to conduct the mission fully autonomously under the guidance of the proposed global planner, as shown in~\cref{fig:exp}. The robot finishes visiting a total of six targets and homing in $4$ minutes and $14$ seconds. The experiment shows that the entire pipeline works in a real-world environment with rough terrain. A single minor collision with a stone happened due to the inaccurate tracking of our collision-free path~(\cref{fig:exp9}).

\section{CONCLUSIONS}
In this work, we present the SMUG planner, a safe multi-goal planner that can rapidly generate an optimal global path and is applicable to various exploration, and industrial inspection missions. With the proposed \gls{idp} algorithm, the planner can generate a safe global path visiting $12$ targets in $8$ seconds and is $8.9$ times faster than the naive approach without losing optimality. The proposed hierarchical state validity checking scheme increases safety, and reduces the total planning time by 30\% compared to pure volumetric collision checking. We conduct experiments in which the ANYmal robot performs a multi-goal mission fully autonomously on rough terrain, showing the real-world applicability of the proposed planner. To the best of our knowledge, this is the first time a global planner is deployed on a mobile robot to a real-world GTSP with collision-free path planning.

We assume the \gls{toi}s can be inspected from all corresponding \gls{poi}s. However, in some real scenarios, if a \gls{toi} is not inspectable from the selected \gls{poi}, a fallback option is needed.

Since the planner is designed for ground robots and uses a robot-specific traversability map, it is not straightforward to generalize the entire method to robots with more degrees of freedom, such as drones. However, \gls{idp} is generalizable to all robots given a corresponding point-to-point planner.


Moreover, the default \gls{ompl} cost for $SE(2)$ space does not capture the traversing capabilities of the legged robot well. 
To counteract this we rely on a simple post-processing heuristic which aligns the waypoints heading to ensure the robot moves forward rather than sideways. A more sensible path cost, eventually learned in simulation, tailored to the motion of the legged robot and a corresponding informed sampler could solve this.










\bibliographystyle{ieeetr}
\bibliography{ref}

\begin{thebibliography}{10}

\bibitem{arches}
A.~Wedler {\em et~al.}, ``Preliminary results for the multi-robot,
  multi-partner, multi-mission, planetary exploration analogue campaign on
  mount etna,'' in {\em 72nd International Astronautical Congress (IAC)},
  October 2021.

\bibitem{esa}
D.~Malaut, ``Esa-esric space resources challenge final,'' October 2022.

\bibitem{hvdc}
C.~Gehring {\em et~al.}, ``Anymal in the field: Solving industrial inspection
  of an offshore hvdc platform with a quadrupedal robot,'' in {\em
  International Symposium on Field and Service Robotics}, 2019.

\bibitem{argos}
{\em {Autonomous Robot for Gas and Oil Sites}}, vol.~All Days of {\em SPE
  Offshore Europe Conference and Exhibition}, 09 2015.

\bibitem{glns}
S.~L. Smith and F.~Imeson, ``{GLNS}: An effective large neighborhood search
  heuristic for the generalized traveling salesman problem,'' {\em Computers \&
  Operations Research}, vol.~87, pp.~1--19, 2017.

\bibitem{tsp}
D.~L. Applegate {\em et~al.}, {\em The Traveling Salesman Problem: A
  Computational Study}.
\newblock Princeton: Princeton University Press, 2007.

\bibitem{rrt*}
S.~Karaman and E.~Frazzoli, ``Sampling-based algorithms for optimal motion
  planning,'' {\em The International Journal of Robotics Research}, vol.~30,
  no.~7, pp.~846--894, 2011.

\bibitem{lvn}
J.~Frey {\em et~al.}, ``Locomotion policy guided traversability learning using
  volumetric representations of complex environments,'' in {\em 2022 IEEE/RSJ
  International Conference on Intelligent Robots and Systems (IROS)},
  pp.~5722--5729, 2022.

\bibitem{rapid}
T.~Cieslewski, , {\em et~al.}, ``Rapid exploration with multi-rotors: A
  frontier selection method for high speed flight,'' in {\em 2017 IEEE/RSJ
  International Conference on Intelligent Robots and Systems (IROS)},
  pp.~2135--2142, 2017.

\bibitem{nbv}
A.~Bircher {\em et~al.}, ``Receding horizon "next-best-view" planner for 3d
  exploration,'' in {\em 2016 IEEE International Conference on Robotics and
  Automation (ICRA)}, pp.~1462--1468, 2016.

\bibitem{gbp}
T.~Dang {\em et~al.}, ``Graph-based path planning for autonomous robotic
  exploration in subterranean environments,'' in {\em 2019 IEEE/RSJ
  International Conference on Intelligent Robots and Systems (IROS)},
  pp.~3105--3112, 2019.

\bibitem{art}
L.~Wellhausen and M.~Hutter, ``Rough terrain navigation for legged robots using
  reachability planning and template learning,'' in {\em 2021 IEEE/RSJ
  International Conference on Intelligent Robots and Systems (IROS)},
  pp.~6914--6921, IEEE Press, 2021.

\bibitem{prm}
L.~Kavraki {\em et~al.}, ``Probabilistic roadmaps for path planning in
  high-dimensional configuration spaces,'' {\em IEEE Transactions on Robotics
  and Automation}, vol.~12, no.~4, pp.~566--580, 1996.

\bibitem{lazyprm*}
K.~Hauser, ``Lazy collision checking in asymptotically-optimal motion
  planning,'' in {\em 2015 IEEE International Conference on Robotics and
  Automation (ICRA)}, pp.~2951--2957, 2015.

\bibitem{informed}
J.~D. Gammell, T.~D. Barfoot, and S.~S. Srinivasa, ``Informed sampling for
  asymptotically optimal path planning,'' {\em IEEE Transactions on Robotics},
  vol.~34, no.~4, pp.~966--984, 2018.

\bibitem{gtsp}
G.~Laporte {\em et~al.}, ``Generalized travelling salesman problem through n
  sets of nodes: An integer programming approach,'' {\em INFOR: Information
  Systems and Operational Research}, vol.~21, no.~1, pp.~61--75, 1983.

\bibitem{tspn}
E.~M. Arkin and R.~Hassin, ``Approximation algorithms for the geometric
  covering salesman problem,'' {\em Discrete Applied Mathematics}, vol.~55,
  no.~3, pp.~197--218, 1994.

\bibitem{ant}
C.-M. Pintea, P.~Pop, and C.~Chira, ``The generalized traveling salesman
  problem solved with ant algorithms,'' {\em J Univers Comput Sci},
  vol.~13(7)rem, 08 2017.

\bibitem{lagrangian}
C.~Noon, ``A lagrangian based approach for the asymmetric generalized traveling
  salesman problem,'' {\em Operations Research}, vol.~39, 08 1991.

\bibitem{transform}
C.~Noon and J.~Bean, ``An efficient transformation of the generalized traveling
  salesman problem,'' {\em INFOR. Information Systems and Operational
  Research}, vol.~31, 02 1993.

\bibitem{antcolony}
J.~Yang, X.~Shi, M.~Marchese, and Y.~Liang, ``Ant colony optimization method
  for generalized tsp problem,'' {\em Progress in Natural Science - PROG NAT
  SCI}, vol.~18, 11 2008.

\bibitem{rkga}
L.~Snyder and M.~Daskin, ``A random-key genetic algorithm for the generalized
  traveling salesman problem,'' {\em European Journal of Operational Research},
  vol.~174, pp.~38--53, 10 2006.

\bibitem{hrkga}
I.~Gentilini, {\em Multi-Goal Path Optimization for Robotic Systems with
  Redundancy based on the Traveling Salesman Problem with Neighborhoods}.
\newblock PhD thesis, Carnegie Mellon University, 2012.

\bibitem{lk}
D.~Karapetyan and G.~Gutin, ``Lin–kernighan heuristic adaptations for the
  generalized traveling salesman problem,'' {\em European Journal of
  Operational Research}, vol.~208, no.~3, pp.~221--232, 2011.

\bibitem{minlp}
I.~Gentilini, F.~Margot, and K.~Shimada, ``The travelling salesman problem with
  neighbourhoods: Minlp solution,'' {\em Optimization Methods and Software},
  vol.~28, no.~2, pp.~364--378, 2013.

\bibitem{onoptimizing}
S.~Alatartsev {\em et~al.}, ``On optimizing a sequence of robotic tasks,'' in
  {\em 2013 IEEE/RSJ International Conference on Intelligent Robots and
  Systems}, pp.~217--223, IEEE, 2013.

\bibitem{mtp}
C.~Wurll, D.~Henrich, and H.~W{\"o}rn, ``Multi-goal path planning for
  industrial robots,'' 1999.

\bibitem{mtpgr}
J.~Faigl, V.~Von{\'a}sek, and L.~Preucil, ``A multi-goal path planning for goal
  regions in the polygonal domain.,'' in {\em ECMR}, pp.~171--176, 2011.

\bibitem{automatic}
W.~Gao {\em et~al.}, ``Automatic task scheduling optimization and
  collision-free path planning for multi-areas problem,'' {\em Intell. Serv.
  Robot.}, vol.~14, p.~583–596, sep 2021.

\bibitem{rba}
X.~Pan {\em et~al.}, ``Approximate shortest path algorithms for sequences of
  pairwise disjoint simple polygons,'' in {\em Proceedings of the 22nd Annual
  Canadian Conference on Computational Geometry, CCCG 2010}, pp.~175--178,
  2010.

\bibitem{gtspn}
K.~Vicencio, B.~Davis, and I.~Gentilini, ``Multi-goal path planning based on
  the generalized traveling salesman problem with neighborhoods,'' in {\em 2014
  IEEE/RSJ International Conference on Intelligent Robots and Systems},
  pp.~2985--2990, 2014.

\bibitem{gbp2}
M.~Kulkarni {\em et~al.}, ``Autonomous teamed exploration of subterranean
  environments using legged and aerial robots,'' in {\em 2022 International
  Conference on Robotics and Automation (ICRA)}, pp.~3306--3313, 2022.

\bibitem{tare}
C.~Cao, H.~Zhu, H.~Choset, and J.~Zhang, ``{TARE: A Hierarchical Framework for
  Efficiently Exploring Complex 3D Environments},'' in {\em Proceedings of
  Robotics: Science and Systems}, (Virtual), July 2021.

\bibitem{voxblox}
H.~Oleynikova {\em et~al.}, ``Voxblox: Incremental 3d euclidean signed distance
  fields for on-board mav planning,'' in {\em 2017 IEEE/RSJ International
  Conference on Intelligent Robots and Systems (IROS)}, pp.~1366--1373, 2017.

\bibitem{ompl}
I.~A. Sucan {\em et~al.}, ``The open motion planning library,'' {\em IEEE
  Robotics \& Automation Magazine}, vol.~19, no.~4, pp.~72--82, 2012.

\bibitem{google}
L.~Perron and V.~Furnon, {\em OR-Tools}.
\newblock Google.

\end{thebibliography}

\end{document}